\documentclass[3p,times,procedia]{elsarticle}
\usepackage{ecrc}
\usepackage[bookmarks=false]{hyperref}
    \hypersetup{colorlinks,
      linkcolor=blue,
      citecolor=blue,
      urlcolor=blue}

\usepackage{nicefrac}       
\usepackage{microtype}      
\usepackage{xcolor}         
\usepackage{graphicx}
\usepackage{amsmath}
\usepackage{amsthm}
\usepackage{booktabs}
\usepackage{times}  
\usepackage{helvet} 
\usepackage{courier}  
\usepackage{tabularx}
\usepackage{amssymb}
\usepackage{bbm}
\usepackage[misc]{ifsym}
\usepackage[export]{adjustbox}
\usepackage{algpseudocode}
\usepackage{algorithmicx}
\usepackage[ruled,vlined,linesnumbered]{algorithm2e}
\usepackage{wrapfig}
\usepackage{subfigure}

\DeclareMathOperator*{\argmin}{arg\,min}

\volume{00}

\firstpage{1}

\journalname{Procedia Computer Science}

\runauth{Author name}

\jid{procs}

\usepackage{amssymb}

\usepackage[figuresright]{rotating}

\begin{document}
\begin{frontmatter}



\dochead{The 14th International Conference on Ambient Systems, Networks and Technologies (ANT) \\ March 15-17, 2023, Leuven, Belgium}%

\title{A Strategy-Oriented Bayesian Soft Actor-Critic Model}


\author[a]{Qin Yang$^*$} 
\author[a]{Ramviyas Parasuraman}

\address[a]{Department of Computer Science, University of Georgia, Athens, GA 30602, USA.}

\begin{abstract}
Adopting reasonable strategies is challenging but crucial for an intelligent agent with limited resources working in hazardous, unstructured, and dynamic environments to improve the system's utility, decrease the overall cost, and increase mission success probability. This paper proposes a novel hierarchical strategy decomposition approach based on the Bayesian chain rule to separate an intricate policy into several simple sub-policies and organize their relationships as Bayesian strategy networks (BSN). We integrate this approach into the state-of-the-art DRL method -- soft actor-critic (SAC) and build the corresponding Bayesian soft actor-critic (BSAC) model by organizing several sub-policies as a joint policy. We compare the proposed BSAC method with the SAC and other state-of-the-art approaches such as TD3, DDPG, and PPO on the standard continuous control benchmarks -- Hopper-v2, Walker2d-v2, and Humanoid-v2 -- in MuJoCo with the OpenAI Gym environment. The results demonstrate that the promising potential of the BSAC method significantly improves training efficiency.
\end{abstract}

\end{frontmatter}

\correspondingauthor[*]{Corresponding author. Tel.: +1-706-308-0481 ; fax: +1-706-542-2966.}
\email{RickYang2014@gmail.com; qy03103@uga.edu}


\enlargethispage{7mm}
\vspace{-10mm}
\section{Introduction}
\label{sec:introduction}

In Artificial Intelligence (AI) methods, a strategy describes the general plan of an AI agent achieving short-term or long-term goals under conditions of uncertainty, which involves setting sub-goals and priorities, determining action sequences to fulfill the tasks, and mobilizing resources to execute the actions \cite{freedman2015strategy}. It exhibits the fundamental properties of agents' perception, reasoning, planning, decision-making, learning, problem-solving, and communication in interaction with dynamic and complex environments \cite{langley2009cognitive,yang2019self}. Especially in the field of real-time strategy (RTS) game \cite{buro2003real,yang2020gut,yang2022game} and real-world implementation scenarios like robot-aided urban search and rescue (USAR) missions \cite{murphy2014disaster,yang2020needs}, agents need to dynamically change the strategies adapting to the current situations based on the environments and their expected utilities or needs \cite{yang2020hierarchical,yang2021can}.

From the agent perspective, a strategy is a rule used by agents to select an action to pursue goals, which is equivalent to a policy in a Markov Decision Process (MDP) \cite{rizk2018decision}. More specially, in reinforcement learning (RL), the policy dictates the actions that the agent takes as a function of its state and the environment, and the goal of the agent is to learn a policy maximizing the expected cumulative rewards in the process. With advancements in deep neural network implementations, deep reinforcement learning (DRL) helps AI agents master more complex strategy (policy) and represents a step toward building autonomous systems with a higher-level understanding of the visual world \cite{arulkumaran2017deep,yang2022self}. 

Although achieving some progress in those domains, DRL is still hard to explain formally how and why the randomization works, which brings the difficulty of designing efficient models expressing the relationships between various strategies (policies) \cite{zhao2020sim}. Especially, it is well-known that the naive distribution of the value function (or the policy representation) across several independent function approximators can lead to convergence problems \cite{matignon2012independent}. 

To address this gap, this paper first introduces the Bayesian Strategy Network (BSN) based on the Bayesian net to decompose a complex strategy or intricate behavior into several simple tactics or actions. 
Then, we propose a new DRL model termed Bayesian Soft Actor-Critic (BSAC), which integrates the Bayesian Strategy Networks (BSN) and the state-of-the-art SAC method. By building several simple sub-policies organizing as BSN, BSAC provides a more flexible and suitable joint policy distribution to adapt to the Q-value distribution, increasing sample efficiency and boosting training performance. We demonstrate the effectiveness of the BSAC against the SAC method and other state-of-the-art approaches, like Policy Optimization (PPO) \cite{schulman2017proximal}, Deep Deterministic Policy Gradient (DDPG) \cite{lillicrap2015continuous}, and Twin Delayed Deep Deterministic (TD3) \cite{fujimoto2018addressing}, on the standard continuous control benchmark domains in MuJoCo \cite{todorov2012mujoco} with the OpenAI Gym \cite{1606.01540} environment. The results show that the promising potential of the BSAC can achieve more efficient sample learning.
\vspace{-3mm}
\section{Background and Preliminaries}
This section provides the essential background about \textit{Bayesian Networks} and \textit{Soft Actor-Critic (SAC)}. When describing a specific method, we use the notations and relative definitions from the corresponding papers.
\vspace{-1mm}
\subsection{Bayesian Networks}
A Bayesian Network structure $\mathcal{G}$ is a directed acyclic graph whose nodes represent random variables $X_1, \cdots, X_n$. Let $Pa_{X_i}^\mathcal{G}$ denote the parents of $X_i$ in $\mathcal{G}$, and NonDescendants$_{X_i}$ denote the variables in the graph that are not descendants of $X_i$. Then $\mathcal{G}$ encodes the following set of conditional independence assumptions, called the local independence, and denoted by $\mathcal{I}_{\ell}(\mathcal{G})$: For each variable $X_i: (X_i \perp NonDescendants_{X_i} \mid Pa_{X_i}^\mathcal{G})$. In other words, the local independence state that each node $X_i$ is conditionally independent of its non-descendants given its parents \cite{koller2009probabilistic}. Furthermore, a Bayesian Net can be presented as the {\it Chain Rule} of conditional probabilities (Eq. \eqref{bn_chain}). Here, $x_i$ is the value of the variable $X_i$, $P(x_i) = P(X_i = x_i)$.
\begin{equation}
\vspace{-5mm}
\begin{split}
    P(x_1 \cap \cdots \cap x_n) = P(x_1)P(x_2 | x_1) \cdots P(x_n | x_1 \cap \cdots \cap x_{n-1})
\label{bn_chain}
\end{split}
\end{equation}

\subsection{Soft Actor-Critic}
 SAC is an off-policy actor-critic algorithm that can be derived from a maximum entropy variant of the policy iteration method. The architecture consider a parameterized state value function $V_\psi(s_t)$, soft Q-function $Q_\theta(s_t, a_t)$, and a tractable policy $\pi_\phi(a_t | s_t)$. It updates the policy parameters by minimizing the Kullback-Leibler divergence between the policy $\pi'$ and the Boltzmann policy in Eq. \eqref{sac_pi}.
\begin{equation}
\vspace{-10mm}
\begin{split}
    \pi_{new} = \argmin_{\pi'} D_{KL} \left( \pi'(\cdot | s_t) \bigg\Vert \frac{exp(Q_\theta(s_t, \cdot)}{Z_\theta(s_t)} \right)
\label{sac_pi}
\end{split}
\end{equation}

\begin{figure}[tbp]
\centering
\includegraphics[width=1\columnwidth]{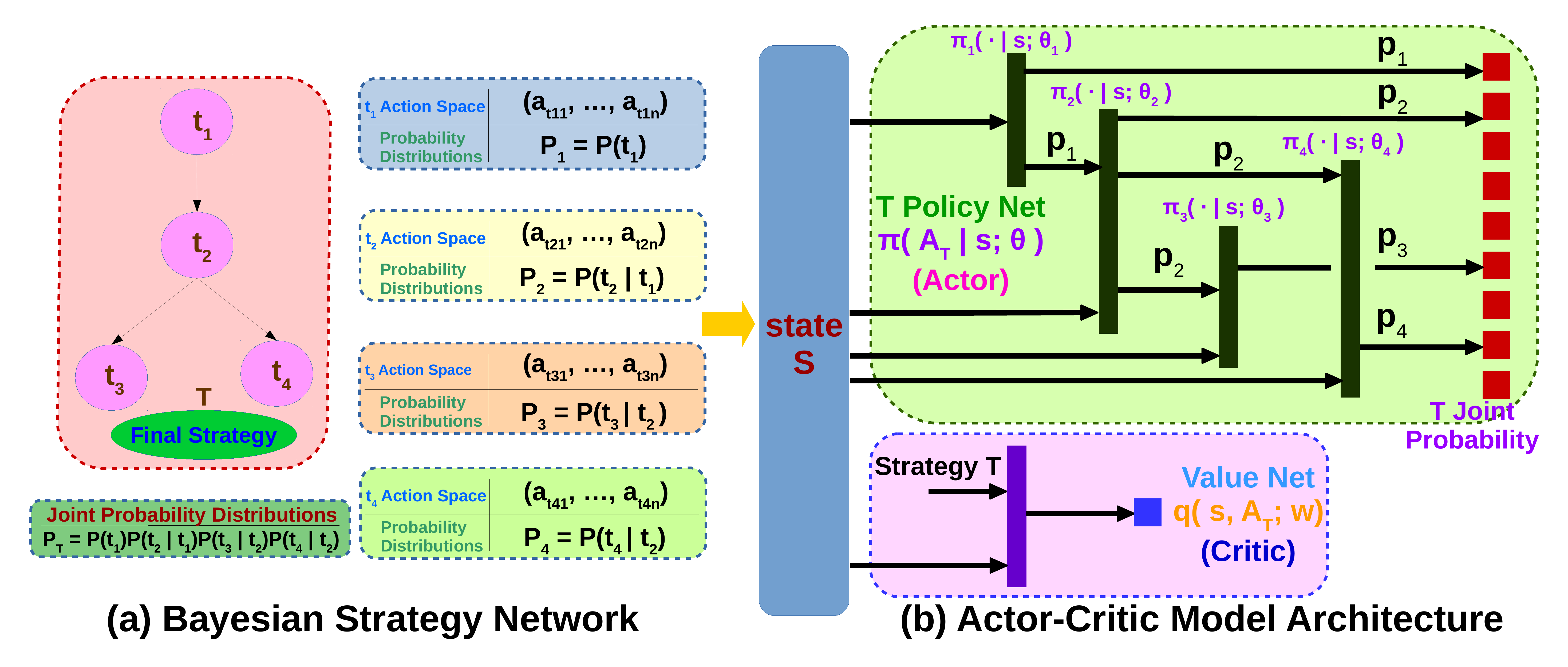}
\vspace{-2mm}
\caption{An overview of the proposed BSN based implementation of the Actor-Critic DRL architecture model.}
\label{bsn_drl}
\vspace{-4mm}
\end{figure}

\section{Methodology}
\label{bsac_method}

Building on top of the well-established suite of actor-critic methods, we introduce the Bayesian Strategy Network (BSN) and integrate the idea of the maximum entropy reinforcement learning framework in SAC, designing the method that results in our Bayesian soft actor-critic (BSAC) approach.

\subsection{Bayesian Strategy Networks (BSN)}
\label{section_bsn}
Supposing that the strategy $\mathcal{T}$ consists of $m$ tactics ($t_1, \dots, t_m$) and their specific relationships can be described as the BSN. We consider the probability distribution $P_{i}$ as the policy for tactic $t_i$. Then, according to the Eq. \eqref{bn_chain}, the joint policy $\pi(a_{\mathcal{T}} \in {\mathcal{T}},s)$ can be described as the joint probability function (Eq. \eqref{agent_task_policy}) through each sub-policy $\pi_i(t_i,s)$, correspondingly. 
An overview of the example BSN implementation in actor-critic architecture is presented in Fig. \ref{bsn_drl}. 
\begin{equation}
\begin{split}
    \pi_{\mathcal{T}}(t_1, \dots, t_m) = \pi_{1}(t_1) \prod_{i=2}^{m} \pi_{i}(t_i | t_1, \dots, t_i),~~~m \in Z^+.
\label{agent_task_policy}
\end{split}
\end{equation}
\subsection{Derivation of Sub-Policy Iteration}
Considering that the agent interacts with an environment through a sequence of observations, strategy (action combinations), and rewards, we can describe the relationships between actions in the strategy as a BSN, represented in Eq. \eqref{agent_task_policy} accordingly. More formally, we can use the corresponding deep convolution neural networks to approximate the strategy \textit{Policy Network} (Actor) in the Eq. \eqref{agent_task_policy} as Eq. \eqref{joint_task_policy} \footnote{Here, $\mathcal{A}$ is the joint action or strategy space for the policy $\pi$; $\theta_i$ and $a_{i_t}$ are the parameters and action space of each sub-policy network $\pi_i$.}. The \textit{Value Network} (Critic) $q(s, \mathcal{A}_t; w)$ evaluate the performance of the specific joint action $\mathcal{A}$ using the value function in Eq. \eqref{joint_task_value} with a parameter $w$. Then, we can calculate the corresponding parameters' gradient descent using Eq. \eqref{gradient_joint_task_policy}.
\vspace{-6mm}
\begin{flalign}
    & \pi(\mathcal{A}_t, s) \approx \pi(\mathcal{A}_t | s; \theta) = \prod_{i=1}^{m} \pi_i(a_{i_t} | s; \theta_i)
\label{joint_task_policy} \\
    & V(s; \theta, w) = \sum_{t \in T}\pi(\mathcal{A}_t | s; \theta) \cdot q(s, \mathcal{A}_t; w)
\label{joint_task_value} \\
    & \frac{\partial V(s; \theta, w)}{\partial \theta}
     = \sum_{i=1}^m \mathop{\mathbb{E}} \left[ \frac{\partial \log \pi_i(a_{i_t} | s; \theta_i)}{\partial \theta} \cdot q(s, \mathcal{A}_t; w) \right]
\label{gradient_joint_task_policy}
\end{flalign}

Through this process, we decompose the strategy policy network $\pi_{\mathcal{T}}$ into several sub-policies networks $\pi_i$ and organize them as the corresponding BSN. Furthermore, according to Eq. \eqref{gradient_joint_task_policy}, each sub-policies uses the same value network to update its parameters in every iteration.

\subsection{Bayesian Soft Actor-Critic (BSAC)}

Our method incorporates the maximum entropy concept into the actor-critic deep RL algorithm. According to the additivity of the entropy, the system's entropy can present as the sum of the entropy of several independent sub-systems \cite{wehrl1978general}. For each step of soft policy iteration, the joint policy $\pi$ will calculate the value to maximize the sum of sub-systems' $\pi_i$ entropy in the BSN using the below objective function Eq. \eqref{bsac_entropy}. In order to simplify the problem, we assume that the weight and the corresponding temperature parameters $\alpha_i$ for each action is the same in each sub-system.
\begin{equation}
\vspace{-3mm}
\begin{split}
    J_V(\pi) = \sum_{t=0}^{T} \mathbb{E}_{(s_t, \mathcal{A}_t) \sim \rho_{\pi_i}} \left [r(s_t, \mathcal{A}_t) + \frac{\alpha}{m} \sum_{i=1}^{m} \mathcal{H}(\pi_i(\cdot|s_t)) \right ]
\label{bsac_entropy}
\end{split}
\end{equation}

The soft Q-value can be computed iteratively, starting from any function $Q: S \times A \rightarrow \mathbb{R}$ and repeatedly applying a modified Bellman backup operator $\mathcal{T}^\pi$ \cite{haarnoja2018soft}. In the Bayesian Soft Actor-Critic, the $\mathcal{T}^\pi$ is given by Eq. \eqref{bsac_bell}. Considering that the evaluation of each sub-policy applies the same Q-value and weight, the soft state value function can be represented in Eq. \eqref{bsac_sqv} \footnote{Here, $Z^{\pi_{old}}(s_t)$ is the partition function to normalize the distribution.}.
\vspace{-3mm}
\begin{equation}
\vspace{-10mm}
\begin{split}
    \mathcal{T}^\pi Q(s_t, \mathcal{A}_t) \triangleq r(s_t, \mathcal{A}_t) + \gamma \mathbb{E}_{s_{t+1} \sim p} \left [V(s_{t+1}) \right]
\label{bsac_bell}
\end{split}
\end{equation}
\begin{equation}
\begin{split}
    V(s_t) = \mathbb{E}_{\mathcal{A}_t \sim \pi_\phi} \left [Q(s_t, \mathcal{A}_t) - \frac{1}{m} \sum_{i=1}^{m} \log \pi_{\phi_i} (a_{i_t} | s_t)) \right]
\label{bsac_sqv}
\end{split}
\end{equation}

Specifically, in each sub-policy $\pi_i$ improvement step, for each state, we update the corresponding policy according to Eq. \eqref{bsac_pi}. 
\vspace{-3mm}
\begin{equation}
\begin{split}
    \pi_{new} = \argmin_{\pi' \in \Pi} D_{KL} \left( \frac{1}{m} \prod_{i=1}^{m} \pi'_{i}(\cdot | s_t) \bigg\Vert \frac{exp(Q^{\pi_{old}}(s_t, \cdot))}{Z^{\pi_{old}}(s_t)} \right)
\label{bsac_pi}
\end{split}
\end{equation}

Furthermore, the soft policy evaluation and the soft policy improvement alternating execution in each soft sub-policy iteration guarantees the convergence of the optimal maximum entropy among the sub-policies combination. We use function approximators for both the Q-function and each sub-policy, optimizing the networks with stochastic gradient descent.
Instead of using one policy network with a Gaussian distribution in SAC to fit the distribution of the Q-value, BSAC generates several simple distributions based on the BSN to adapt to the given model. 
\vspace{-3mm}
\section{Evaluation}
We evaluate the performance of the proposed BSAC agent in several challenging continuous control environments with varying action combinations and complexity. We choose the MuJoCo physics engine \cite{todorov2012mujoco} to simulate our experiments in the OpenAI's Gym environment \cite{1606.01540}. In our experiments, we use three of the standard continuous control benchmark domains -- Hopper-v2, Walker2d-v2, and Humanoid-v2.
We study the performance of the proposed BSAC against the state-of-the-art continuous control algorithm, the SAC \cite{haarnoja2018soft} and other benchmark DRL algorithms, PPO \cite{schulman2017proximal}, DDPG \cite{lillicrap2015continuous}, and TD3 \cite{fujimoto2018addressing}.

\begin{figure*}[t]
\centering
\begin{minipage}[b]{0.265\linewidth}
\includegraphics[width=1\textwidth]{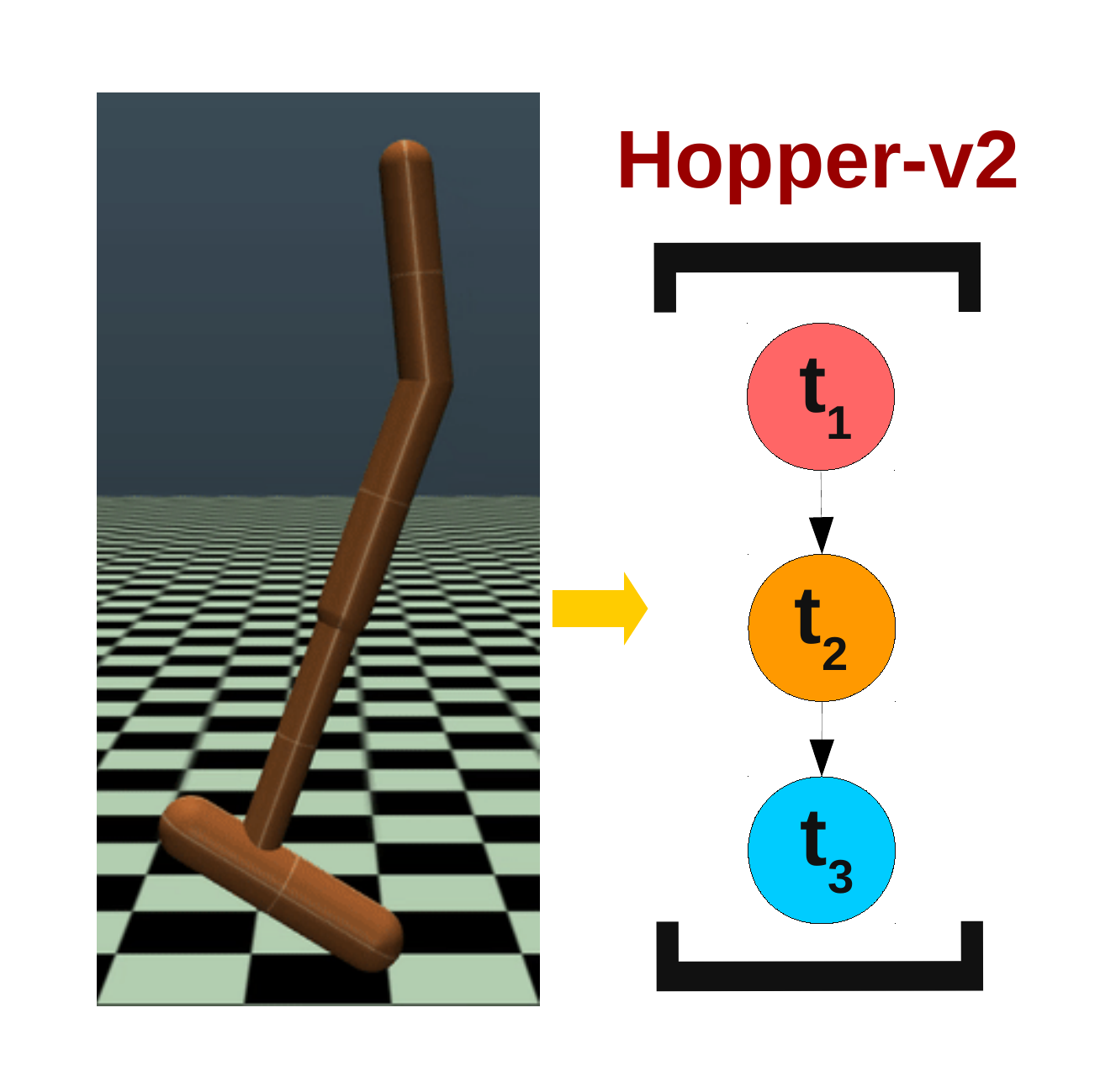}\vspace{0pt}
\caption{\small{BSN 3P Model in Hopper-v2}}
\label{hopper_2v_bsn}
\end{minipage}
\begin{minipage}[b]{0.314\linewidth}
\includegraphics[width=1\textwidth]{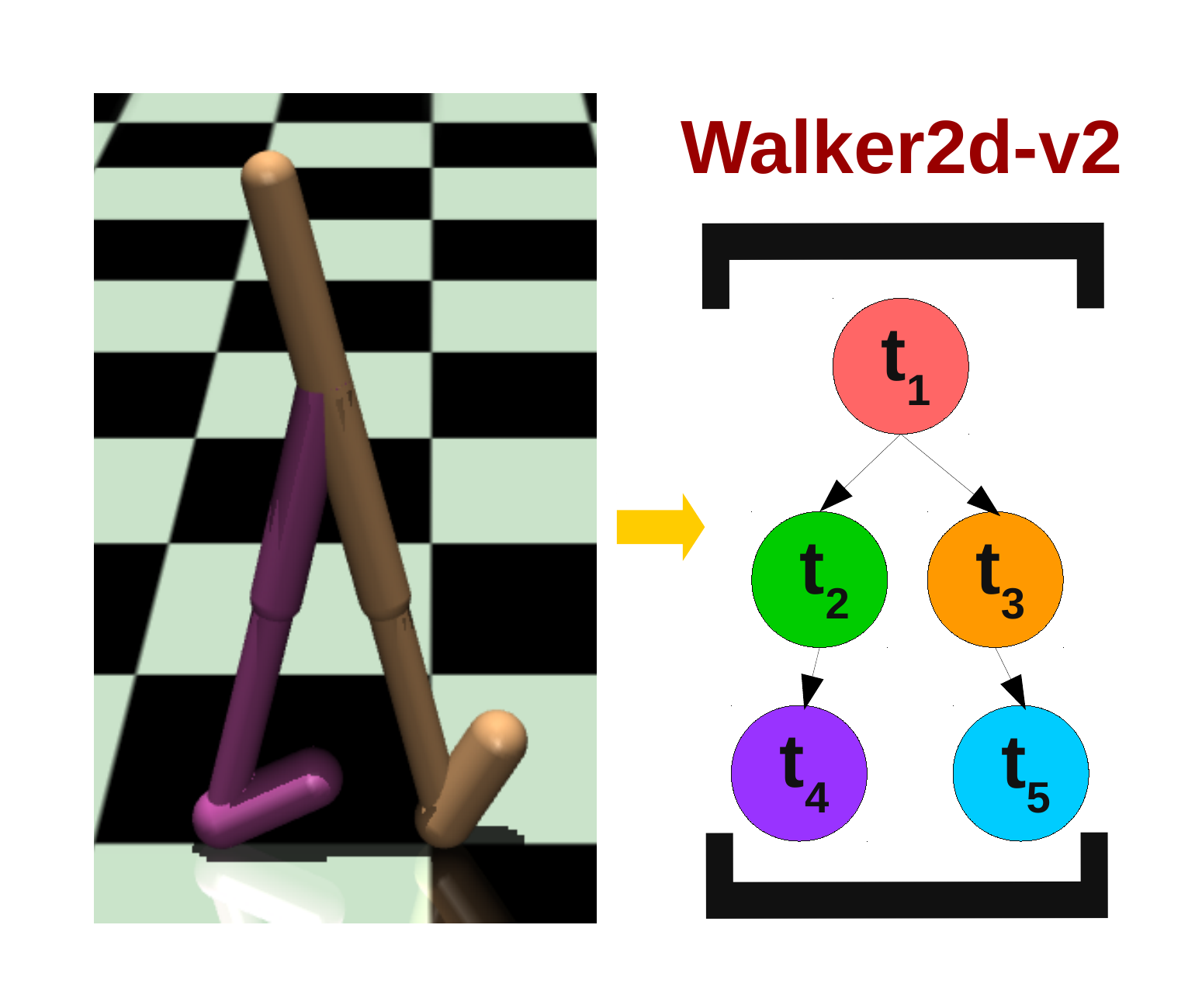}\vspace{0pt}
\caption{\small{BSN 5P Model in Walker2d-v2}}
\label{walker2d_2v_bsn}
\end{minipage}
\begin{minipage}[b]{0.38\linewidth}
\includegraphics[width=1\textwidth]{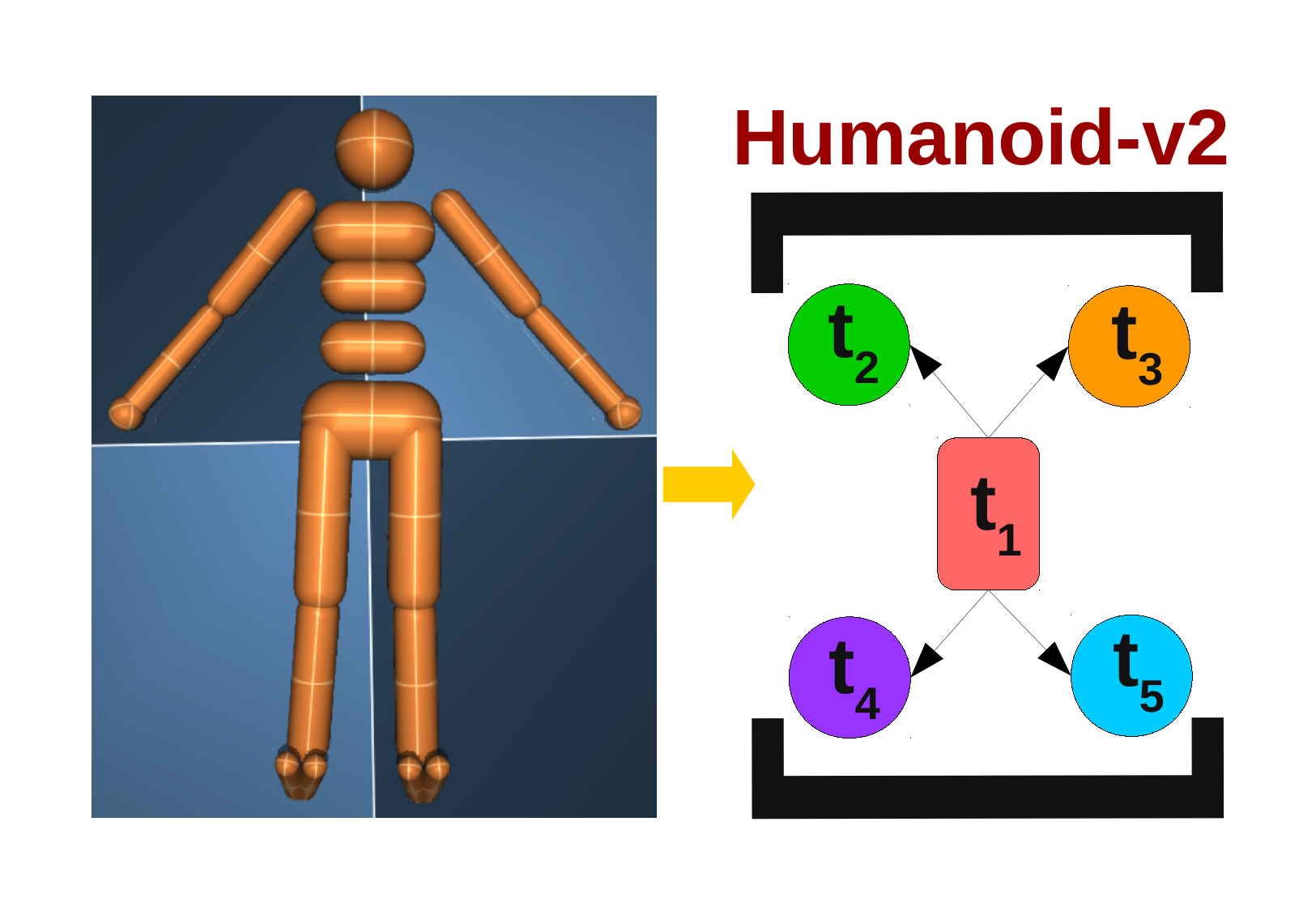}\vspace{0pt}
\caption{\small{BSN 5P Model in Humanoid-v2}}
\label{fig:humanoid_5bsn}
\end{minipage}
 \vspace{-6mm}
\end{figure*}
\begin{figure*}[t]
\centering
    \subfigure[Hopper-v2 Domain]{
    \begin{minipage}[t]{0.32\linewidth}
 \includegraphics[width=1\textwidth]{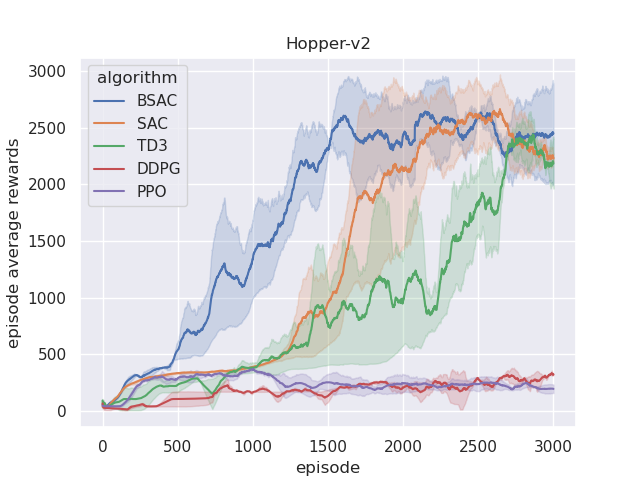}
 \vspace{-2mm}
 \label{fig: hopper-v2}
 \end{minipage}}
 \subfigure[Walker2d-v2 Domain]{
    \begin{minipage}[t]{0.32\linewidth}
 \includegraphics[width=1\textwidth]{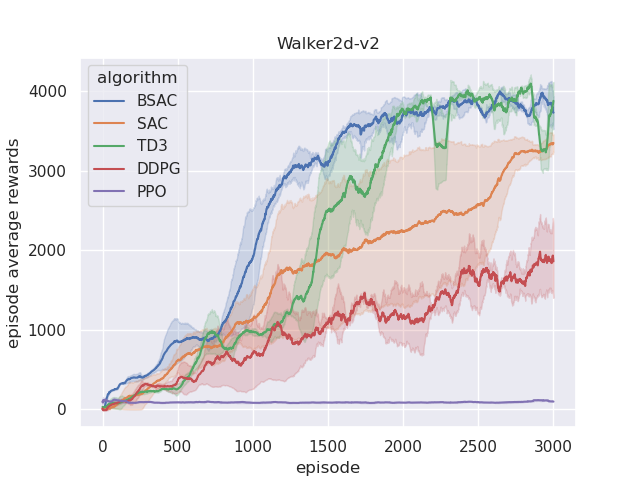}
 \vspace{-2mm}
 \label{fig: walker2d-v2}
 \end{minipage}}
 \subfigure[Humanoid-v2 Domain]{
    \begin{minipage}[t]{0.32\linewidth}
    \includegraphics[width=1\textwidth]{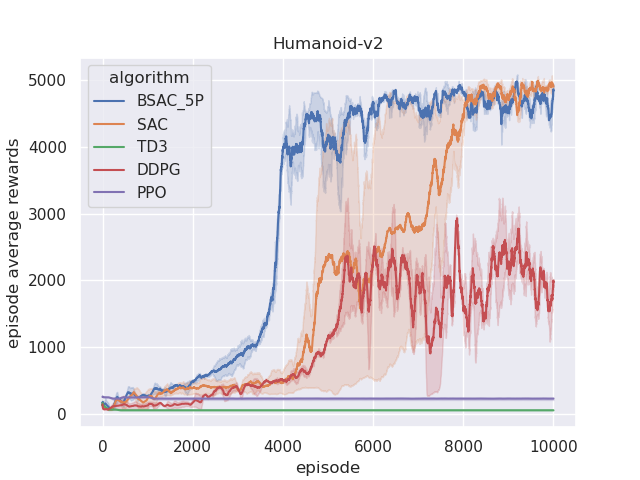}
    \vspace{-2mm}
    \label{fig: humanoid-v2}
    \end{minipage}}
    \vspace{-2mm}
\caption{\small{The Performance Results in the OpenAI Gym with MuJoCo physics engine.}}
\label{fig: mojuco_experiment}
\vspace{-4mm}
\end{figure*}

In the corresponding experiments, we decompose hopper, walker2d, and humanoid behaviors into different BSNs. Such as three policies (3P) BSN (hip action $t_1$, knee action $t_2$, and ankle action $t_3$) in Hooper-v2 (Fig. \ref{hopper_2v_bsn}), five policies (5P) BSN (hip action $t_1$, left knee action $t_2$, right knee action $t_3$, left ankle action $t_4$, and right ankle action $t_5$) in Walker2d-v2 (Fig. \ref{walker2d_2v_bsn}), and five policies (5P) BSN (abdomen action $t_1$, the actions $t_2$ of right hip and right knee, the actions $t_3$ of left hip and left knee, the actions $t_4$ of right shoulder and right elbow, and actions $t_5$ of left shoulder and left elbow) in Humanoid-v2 (Fig.~\ref{hopper_2v_bsn}). Furthermore, for corresponding BSAC models, we can formalize their joint policy (action) as Eq. \eqref{hopper_bsn}, Eq. \eqref{walker_bsn}, and Eq. \eqref{humanoid_bsn}.
\vspace{-6mm}
\begin{equation}
\begin{split}
    P(t_1, t_2, t_3) = P(t_1) P(t_2|t_1) P(t_3|t_2)
\label{hopper_bsn}
\end{split}
\end{equation}
\vspace{-16mm}
\begin{equation}
\begin{split}
    P(t_1, t_2, t_3, t_4, t_5) = P(t_1)P(t_2|t_1)P(t_3|t_1)P(t_4|t_2)P(t_5|t_3)
\label{walker_bsn}
\end{split}
\end{equation}
\vspace{-16mm}
\begin{equation}
\begin{split}
    P(t_1, t_2, t_3, t_4, t_5) = P(t_1)P(t_2|t_1)P(t_3|t_1)P(t_4|t_1)P(t_5|t_1)
\label{humanoid_bsn}
\end{split}
\end{equation}

Comparing the performance of the BSAC with SAC, TD3, DDPG, and PPO in Hopper-v2 (Fig. \ref{fig: hopper-v2}), Walker2d-v2 (Fig. \ref{fig: walker2d-v2}), and Humanoid-v2 (Fig. \ref{fig: humanoid-v2}), we prove that BSAC can achieve higher performance than other DRL algorithms. Furthermore, with the increasing complexity of the agent's behaviors and strategy, decomposing the complex behaviors into simple actions or tactics and organizing them as a suitable BSN, building the corresponding joint policy model in the BSAC can substantially increase training efficiency. Furthermore, we plan to open the source of the BSAC algorithm on GitHub: \url{https://github.com/RickYang2016/Bayesian-Soft-Actor-Critic--BSAC}.

Generally speaking, the BSAC provides an approach to generating a more suitable joint policy to fit the value distribution, improving the convergence efficiency and the model's performance. Specifically, a conventional RL approach is to specify a unimodal policy distribution centered at the maximal Q-value and extend to the neighboring actions to provide noise for exploration \cite{haoren2017}. Especially the exploration is biased toward the local passage, the agent refines its policy there and ignores others completely \cite{haarnoja2017reinforcement}. In other words, if we can design a suitable joint policy distribution consisting of several simple policy distributions to fit the corresponding Q-value distribution, it will essentially boost the sample efficiency in the agent's training.
\vspace{-3mm}
\section{Conclusions}
We introduce a novel agent strategy composition approach termed Bayesian Strategy Network (BSN) for achieving efficient deep reinforcement learning (DRL). Through integrating the BSN and SAC, we propose the Bayesian soft actor-critic (BSAC) approach to decompose an intricate strategy or joint policy into several simple sub-policies and organize them as the knowledge graph.
We demonstrate it on the standard continuous control benchmark, such as the Hopper, Walker, and the Humanoid, in MuJoCo with the OpenAI Gym environment. The results demonstrate the potential and significance of the proposed BSAC architecture by achieving more efficient sample learning and higher performance against the state-of-the-art DRL methods. Furthermore, implementing the BSAC on real robots is also an interesting problem. It will help us develop robust computation models for robotic systems, such as robot locomotion control, multi-robot planning and navigation, and robot-aided search and rescue missions.
\vspace{-3mm}
\bibliographystyle{acm}
\bibliography{references}

\end{document}